\documentclass[letterpaper, 10 pt, conference]{ieeeconf}  

\IEEEoverridecommandlockouts                              

\usepackage{graphicx} 
\usepackage{epsfig} 
\usepackage{times} 
\usepackage{amsmath} 
\usepackage{amssymb}  
\usepackage{cite}
\usepackage{multirow}
\usepackage{multicol}
\usepackage{array}

\UseRawInputEncoding

\title{\LARGE \bf
Improved Radar Localization on Lidar Maps Using Shared Embedding
}

\author{Huan Yin, Yue Wang and Rong Xiong
\thanks{The authors are with the State Key Laboratory of Industrial Control Technology and Institute of Cyber-Systems and Control, Zhejiang University, Zhejiang, China.}%
}

\begin{document}

\maketitle
\thispagestyle{empty}
\pagestyle{empty}

\begin{abstract}
We present a heterogeneous localization framework for solving radar global localization and pose tracking on pre-built lidar maps. To bridge the gap of sensing modalities, deep neural networks are constructed to create shared embedding space for radar scans and lidar maps. Herein learned feature embeddings are supportive for similarity measurement, thus improving map retrieval and data matching respectively. In RobotCar and MulRan datasets, we demonstrate the effectiveness of the proposed framework with the comparison to Scan Context and RaLL. In addition, the proposed pose tracking pipeline is with less neural networks compared to the original RaLL.

\end{abstract}

\begin{figure*}[t]
	\centering
	\includegraphics[width=18cm]{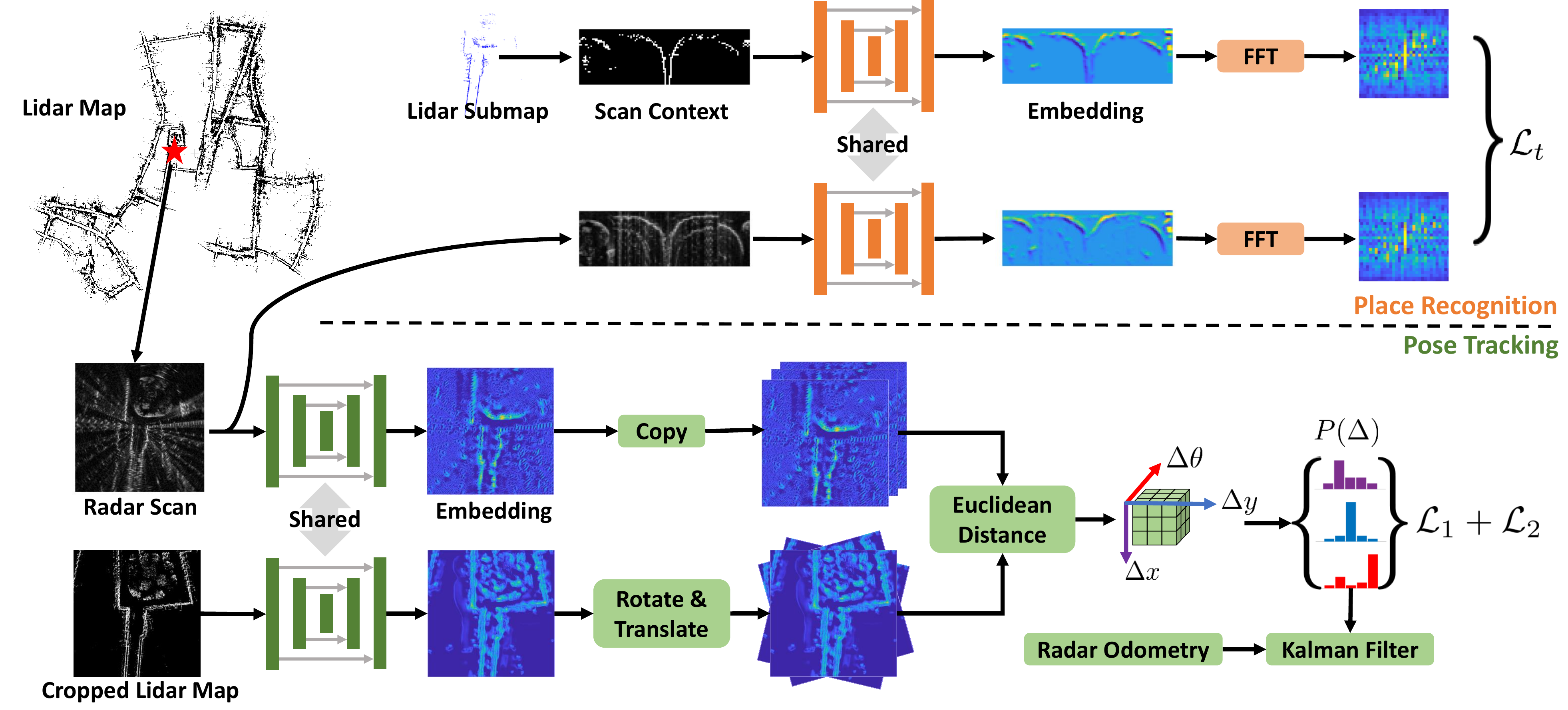}
	\caption{Our proposed framework for radar localization on pre-built lidar maps.}
	\label{methods}	
\end{figure*}

\section{Introduction}

Localization on map aims at estimating robot pose with sensor measurements and pre-built maps, which is fundamental for robot self-awareness. However, in adverse conditions, visual and lidar based localization suffer a lot from illumination changes and extreme weathers. Recent radar sensing brings robust perception for mobile platforms \cite{barnes2020oxford,kim2020mulran,sheeny2020radiate}, thus we consider that radar localization on map is a good solution to overcome environmental challenges.

Since large-scale high-resolution lidar maps have been widely deployed on roads, we follow our previous method RaLL \cite{yin2021rall} and propose to localize the radar sensor on lidar maps. Figures~\ref{methods} presents our proposed lidar maps-aided radar localization framework, which consists two pipelines: \textit{place recognition} without prior pose and \textit{pose tracking} for online precise localization. Both two pipelines use siamese U-Net to create shared features emebddings for similarity measure or data matching.


\section{Methods}

\subsection{Radar-to-lidar place recognition}

We first present our proposed radar-to-lidar place recognition method. We borrow an idea from our previous research work DiSCO \cite{xu2021disco}, in which learned Scan Context \cite{kim2018scan} representation is utilized for lidar place recognition and orientation estimation. 

To achieve radar-to-lidar place recognition, the original DiSCO is updated from two aspects. Firstly, lidar submaps are constructed by accumulating laser scans, which are more applicable for heterogenous place recognition. Secondly, we propose to train the neural network with all mixed combinations and a triplet loss for joint learning. The designed triplet loss is formulated as follows:
\begin{equation} \label{loss}
	\mathcal{L}_{t} = \frac{1}{\left| \mathcal{F} \right|} \sum_{F\in\mathcal{F}} \text{max}(0, m + \text{pos}(F) - \text{neg}(F))
\end{equation}
where $m$ is a margin value and the set $\mathcal{F}$ contains all combinations for \{anchor, positive, negative\}, such as \{radar, radar, lidar\} and \{lidar, radar, radar\} etc. There are $2^3$ candidates for every three places. $\text{pos}(F)$ and $\text{neg}(F)$ are measured Euclidean distances of a pair of postives and a pair of negatives respectively.

\subsection{Radar-on-lidar pose tracking}

In our preivous work \cite{yin2021rall}, we proposed an end-to-end radar localization framework. To improve the localization accuracy and efficiency, we first utilize one siamese network (U-Net) to extract shared embeddings rather than separate extraction modules in \cite{yin2021rall}. Additionally, we use Euclidean distances to measure the similarities between transformed map candidates and radar data, rather than complex patch network in \cite{yin2021rall}. Overall, the complicated networks of RaLL is simplified to a more concise form, as shown in Figure~\ref{methods}.

We train the feature extraction module with a cross entropy loss $\mathcal{L}_1$ and squared errors $\mathcal{L}_2$, as follows:
\begin{equation}\label{L_1}
	\mathcal{L}_1 =-\sum_{\Delta x_i}C_x\log(P_x)
	- \sum_{\Delta y_i} C_y\log(P_y)
	-\sum_{\Delta \theta_k} C_\theta\log(P_\theta)
\end{equation}
\begin{equation}\label{L_2}
	\mathcal{L}_2 = (\Delta \hat{x} - \Delta x)^2 + (\Delta \hat{y} - \Delta y)^2 + \alpha \cdot (\Delta \hat{\theta} - \Delta \theta)^2
\end{equation}
in which $\{ C_x,C_y,C_\theta \}$ and $\{ \Delta x,\Delta y,\Delta \theta \}$ are with ground truth of pose offset, while $\{ P_x,P_y,P_\theta \}$ and $\{ \Delta \hat{x},\Delta \hat{y},\Delta \hat{\theta} \}$ are with estimated offsets. We train the three-layer U-Net with $\mathcal{L}_1 + \mathcal{L}_2$ directly.

\section{Experimental Results}

We evaluate our proposed methods with two public datasets, RobotCar and MulRan. The proposed two network structures are first trained with part of one sequence in RobotCar, and then generalized to other sequences of RobotCar and unseen scenes in MulRan without retraining.

\subsection{Place recognition}

Our joint learning based DiSCO is compared to Scan Context \cite{kim2018scan, kim2020mulran}. For global localization task, we perform multi-session place recognition, which means that map data and query data are generated from different sequences at same place. The source code is released online \footnote{https://github.com/ZJUYH/radar-to-lidar-place-recognition}.

The evaluation results are shown in Table~\ref{recall}, and ``R2R'' stands for radar-to-radar place recognition for instance. Note that we set distance threshold as 3m for evaluation. Our proposed learning based method generally outperforms than the original Scan Context method, thus validating the superiority of learned shared embeddings.

\begin{table}[!t]
	\begin{center}
		\renewcommand\arraystretch{1.0}
		\caption{Recall@1 (\%) of multi-session place recognition}
		\label{recall}
		\begin{tabular}{p{1.8cm}<{\centering}p{1cm}<{\centering}|p{2cm}<{\centering}p{2.2cm}<{\centering}}
			\hline
			\hline
			\multicolumn{2}{c|}{Dataset} & Scan Context \cite{kim2018scan}& Shared Embedding \\
			\hline
			\multirow{3}*{RobotCar-Test} & L2L & 92.51 & \textbf{93.68} \\
			~ & R2R & 91.35 & \textbf{93.18} \\
			~ & R2L & 1.35 & \textbf{61.23} \\
			\hline
			\multirow{3}*{MulRan-Riverside} & L2L & 37.22 & \textbf{38.89} \\
			~ & R2R & \textbf{42.68} & 41.15 \\
			~ & R2L & 1.17 & \textbf{26.20} \\
			\hline
			\multirow{3}*{MulRan-KAIST} & L2L & 28.81 & \textbf{65.01} \\
			~ & R2R & \textbf{72.65} & 66.56 \\
			~ & R2L & 0.55 & \textbf{63.22} \\
			\hline
			\hline
		\end{tabular}
	\end{center}
\end{table}

\subsection{Pose tracking}

We follow the experimental settings in \cite{yin2021rall} \footnote{https://github.com/ZJUYH/RaLL}. The pose tracking evaluation is presented in Table~\ref{tracking}. RaLL performs better at translational estimation, while the proposed method is with less rotational errors. Overall, the improved pose tracking method using shared embedding also achieves comparable performance, even with a simplified structure and less neural networks.

\begin{table}[!t]
	\begin{center}
		\renewcommand\arraystretch{1.0}
		\caption{RMSE of pose tracking}
		\label{tracking}
		\begin{tabular}{p{2.0cm}<{\centering}|p{1.0cm}<{\centering}p{1.0cm}<{\centering}|p{1.0cm}<{\centering}p{1.0cm}<{\centering}}
			\hline
			\hline
			\multirow{2}*{Dataset} & \multicolumn{2}{c|}{RaLL \cite{yin2021rall}}& \multicolumn{2}{c}{Shared Embedding} \\
			~ & Trans.(m) & Rot.($^\circ$) & Trans.(m) & Rot.($^\circ$) \\
			\hline
			Oxford-04 & \textbf{1.71} & 1.93 & 2.17 & \textbf{1.88}\\
			Oxford-05 & \textbf{1.11} & \textbf{1.48} & 1.48 & 1.50 \\
			Oxford-06 & \textbf{1.14} & 1.52 & 1.56 & \textbf{1.50} \\
			DCC-01 & \textbf{2.11} & 1.97 & 2.66& \textbf{1.47} \\
			DCC-02 & \textbf{4.71} & 2.01 & 4.76& \textbf{1.19} \\
			DCC-03 & \textbf{5.14} & 2.55 & 5.86& \textbf{2.36} \\
			KAIST-01 & \textbf{1.45} & 1.74 & 1.89 & \textbf{1.18} \\
			KAIST-02 & \textbf{1.30} & 1.71& 1.80 & \textbf{1.08}\\
			KAIST-03 & \textbf{1.27} & 1.50 & 1.91 & \textbf{1.04}\\
			Riverside-01 & \textbf{4.12} & 2.84 & 4.41& \textbf{1.38}\\
			Riverside-02 & 2.52 & 1.93 & \textbf{2.51} & \textbf{1.29} \\
			\hline
			\hline
		\end{tabular}
	\end{center}
\end{table}

\section{Conclusion}

The shared embedding-based framework not ony solves the heterogenous place recognition for large-scale global localization, but also tracks the mobile robot without failure over 90km driving. With quantitative experiments, our proposed method improves radar localization performance with a more simiplified approach. 

Furthermore, we consider that only one network might be desired for radar based global localization and pose tracking, thus making the radar localization module more consice for real applications.

\addtolength{\textheight}{-12cm}   

\bibliographystyle{IEEEtran}
\bibliography{root}

\end{document}